%% file: main.tex
\documentclass[10pt,twocolumn,letterpaper]{article}

\usepackage[pagenumbers]{cvpr} 


\definecolor{cvprblue}{rgb}{0.21,0.49,0.74}
\usepackage[pagebackref,breaklinks,colorlinks,allcolors=cvprblue]{hyperref}

\usepackage{multirow}
\usepackage[accsupp]{axessibility}  

\def\paperID{*****} 
\def\confName{CVPR}
\def\confYear{2025}

\usepackage[capitalize]{cleveref}
\crefname{section}{Sec.}{Secs.}
\Crefname{section}{Section}{Sections}
\Crefname{table}{Table}{Tables}
\crefname{table}{Tab.}{Tabs.}

\title{HumMorph: Generalized Dynamic Human Neural Fields from Few Views}

\author{Jakub Zadrożny \quad Hakan Bilen\\[2pt]
University of Edinburgh\\[2pt]
{\tt\small \url{https://jakubzadrozny.github.io/hummorph}}
}

\begin{document}

\twocolumn[{%
\renewcommand\twocolumn[1][]{#1}%
\maketitle
\begin{center}
    \centering
    \captionsetup{type=figure}
    \includegraphics[width=0.99\linewidth]{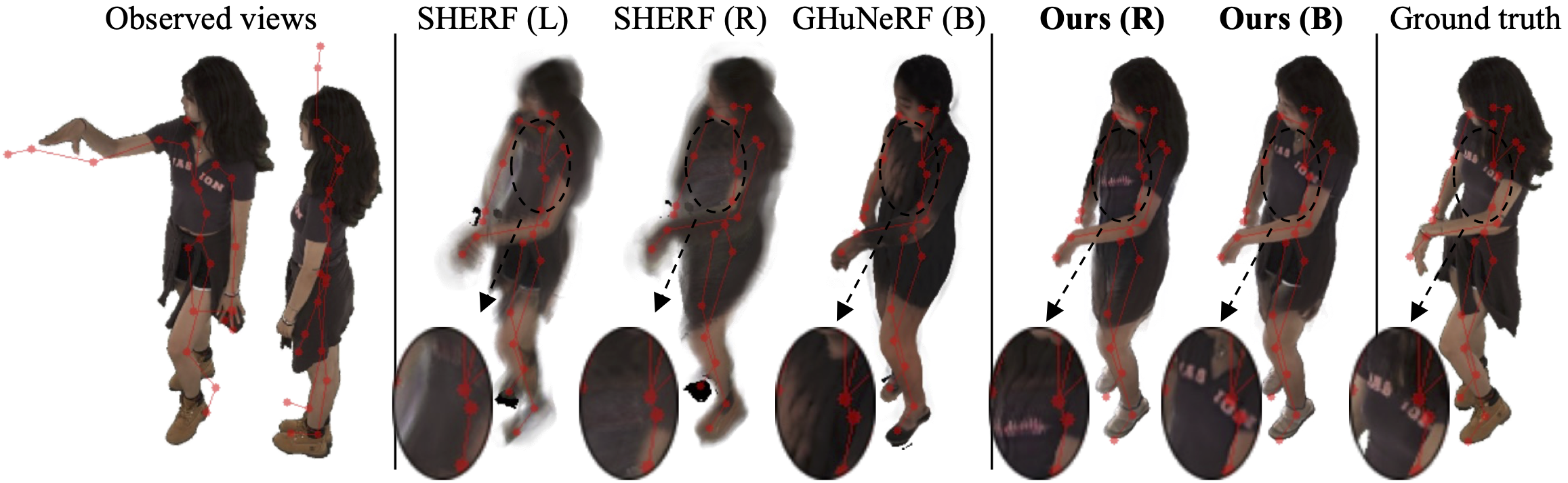}
    \captionof{figure}{\textbf{HumMorph} is a generalized method for free-viewpoint synthesis of humans in novel poses given a few observations. 
    State-of-the-art methods,
    including SHERF \cite{hu_sherf_2023} and GHuNeRF \cite{li_ghunerf_2024}, require accurate body pose annotations for the observed views. These are typically unavailable in practice and the poses need to be noisily estimated instead, like in the example above (poses shown in red). In this scenario, existing approaches struggle to model details and synthesize oversmoothed renders. In contrast, our approach includes dense 3D processing modules and accounts for pose estimation errors to accurately recover detail.
    The letter in parentheses indicates which views were supplied to the methods: L -- left, R -- right or B -- both.}
    \label{fig:cover}
\end{center}%
}]

\input{sec/0_abstract}

\input{sec/1_intro}
\input{sec/2_related}
\input{sec/3_method}
\input{sec/4_experiments}
\input{sec/5_conclusion}

{
    \small
    \bibliographystyle{ieeenat_fullname}
    \bibliography{main}
}

\input{sec/X_suppl}

\end{document}

%% file: sec/0_abstract.tex
\begin{abstract}
We introduce HumMorph, a novel generalized approach to free-viewpoint rendering of dynamic human bodies with explicit pose control. HumMorph renders a human actor in any specified pose given a few observed views (starting from just one) in arbitrary poses.
Our method enables fast inference as it relies only on feed-forward passes through the model.
We first construct a coarse representation of the actor in the canonical T-pose, which combines visual features from individual partial observations and fills missing information using learned prior knowledge.
The coarse representation is complemented by fine-grained pixel-aligned features extracted directly from the observed views, which provide high-resolution appearance information.
We show that HumMorph is competitive with the state-of-the-art when only a single input view is available, however, we achieve results with significantly better visual quality given just 2 monocular observations. 
Moreover, previous generalized methods assume access to accurate body shape and pose parameters obtained using synchronized multi-camera setups.
In contrast, we consider a more practical scenario where these body parameters are noisily estimated directly from the observed views. Our experimental results demonstrate that our architecture is more robust to errors in the noisy parameters and clearly outperforms 
the state of the art in this setting.
\end{abstract}

%% file: sec/1_intro.tex
\section{Introduction}
\label{sec:intro}
Effortless and efficient synthesis of high-quality, realistic humans in previously unseen poses is essential for building a realistic and vibrant Metaverse. It has natural applications directly related to augmented/virtual reality (AR/VR), such as 3D immersive communication, but also in wider content creation including movie production.
In this work, we focus on learning models that can synthesize humans solely from monocular frames without requiring costly multi-camera capturing setups.
This is a key step towards in-the-wild applicability, as generally only non-specialized capturing equipment, such as mobile devices, is available. 

Despite the remarkable progress in human modeling from monocular videos, most approaches~\cite{peng_animatable_2021, wang_arah_2022, li_tava_2022, weng_humannerf_2022, yu_monohuman_2023, guo_vid2avatar_2023} require training a separate model for each subject, which heavily limits their applicability in practice due to compute and energy requirements.
HumanNeRF~\cite{weng_humannerf_2022} maps all observations to a canonical Neural Radiance Field (NeRF)~\cite{mildenhall_nerf_2022} and learns a motion field mapping from observation to canonical space.
However, subject-specific models such as HumanNeRF~\cite{weng_humannerf_2022} require extensive observations for each subject and fail to in-paint details that are not visible in the observations, as they do not capture and inject prior from multiple subjects.

Recent works~\cite{hu_sherf_2023, li_ghunerf_2024, masuda_generalizable_2024, dey_ghnerf_2024, pan_transhuman_2023} learn `generalized' human models from multiple identities and generalize to a previously unseen target identity and their unseen poses from a set of observations in a single forward pass, which significantly speeds up inference and makes them more suitable for real-world applications.
However, SHERF~\cite{hu_sherf_2023} is designed to learn only from a single view of a human actor and cannot combine multiple observations to provide high-quality synthesis even when multiple views are available.
In contrast, GHuNeRF~\cite{li_ghunerf_2024} allows aggregating information from multiple frames of a monocular video, however, to resolve occlusions, it relies on pre-computed visibility masks of a template body mesh. Such masks are not always accurate, especially when body pose is estimated from sparse observed monocular views, and GHuNeRF \cite{li_ghunerf_2024} lacks a mechanism to account for that.
Both SHERF and GHuNeRF heavily rely on deformations of an appropriate SMPL \cite{loper_smpl_2015} mesh to spatially register points across different body poses.
In the common human datasets~\cite{cai_humman_2022, cheng_dna-rendering_2023} used for evaluations, the SMPL parameters are accurately estimated from synchronized multi-view camera setups, which in practice are not available.
These parameters should instead be estimated directly from the observed views, resulting in more noisy estimations and significantly worse reconstructions, as shown in \cref{fig:cover} and our experimental results.

To tackle this challenge, we propose \textit{HumMorph}, a novel efficient generalized model that can synthesize subjects effortlessly from several frames of a monocular video without relying on a template body mesh and using only feed-forward network passes.
Our model incorporates prediction of skinning weights, which eliminates the requirement for accurate body shape parameters and leads to more robust human synthesis.
At the heart of our architecture lies our \textit{VoluMorph} module that 
lifts the observed views from 2D to 3D, then aligns them to the canonical pose (T-pose), and finally combines them through 3D convolutions and attention-based aggregation.
The combined feature volumes are used to estimate a coarse body model in the canonical pose and a residual correction to initial heuristic skinning weights.
Similarly to existing generalized approaches, we require the body pose parameters, but we do not require the body shape parameters and instead rely on the 3D skeleton shape in the canonical T-pose.
Given accurate poses, we demonstrate results of state-of-the-art perceptual visual quality using a single observed view with a significant boost in quality when two input views are available and further improvements with additional observations.
Crucially, our experiments show that our architecture, thanks to using dense 3D processing, is significantly more robust than the state of the art when provided with body pose parameters noisily estimated from the observed views.

\begin{figure*}
    \centering
    \includegraphics[width=\linewidth]{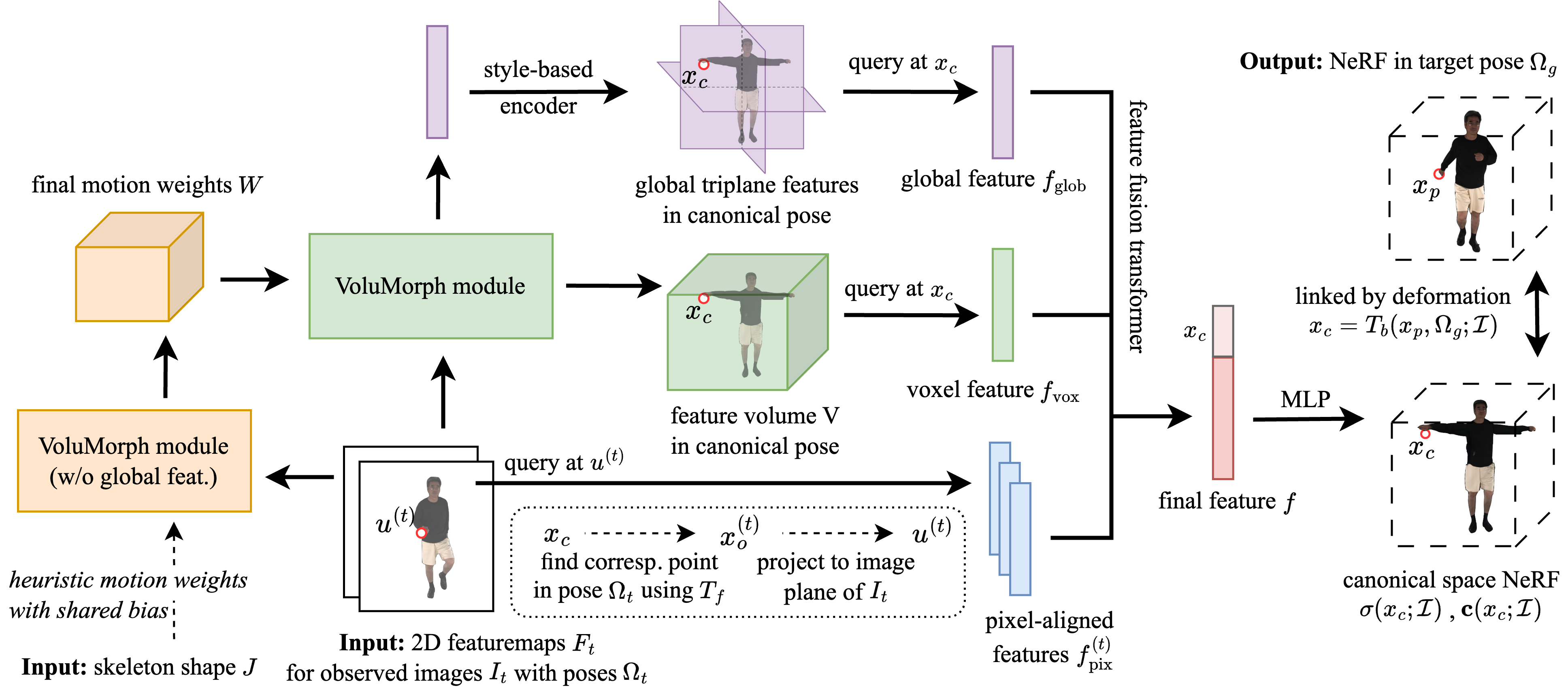}
    \caption{An overview of our approach. First, we extract the 2D featuremaps $F_t$, which we pass through a VoluMorph module to get the final motion weights $W$.
    The features $F_t$ and motion weights $W$ are passed to a second VoluMorph module, which outputs the volume $V$ and a global latent code. Finally, we extract $f_\textrm{vox}, f_\textrm{glob}, f_\textrm{pix}$ and combine them using the feature fusion module to condition the NeRF MLP.}
    \label{fig:method-overview}
\end{figure*}

%% file: sec/2_related.tex
\section{Related Work}
\label{sec:related-work}

\textbf{Neural scene representations.} Our work builds on the recent advancements in neural rendering techniques. NeRF \cite{mildenhall_nerf_2022} assigns density and color values to points in the 3D space, which enables novel view rendering via alpha-compositing densities and colors along camera rays. The original NeRF~\cite{mildenhall_nerf_2022} requires per-scene optimization and a dense set of observed views for supervision. Multiple conditional variants of NeRF were proposed \cite{yu_pixelnerf_2021, lin_vision_2023, henzler_unsupervised_2021, wang_ibrnet_2021, reizenstein_common_2021} to enable feed-forward inference given sparse observed views. However, these only target \textit{static} scenes.

\textbf{Neural fields for humans.}
The use of parametric body models \cite{loper_smpl_2015, pavlakos_expressive_2019, alldieck_imghum_2021}  enabled learning dynamic neural fields for humans by providing a prior for the geometry and an accurate deformation model. The dominant approach \cite{peng_animatable_2021, wang_arah_2022, li_tava_2022, weng_humannerf_2022, jiang2022neuman, yu_monohuman_2023, guo_vid2avatar_2023, liu_hosnerf_2023}
is to model the body in the canonical pose and, during volumetric rendering, deform the query points from observation space to the canonical space using linear blend skinning (LBS). The LBS deformation, however, cannot accurately capture soft elements like muscles, hair, or clothing. 
To address this issue, several approaches \cite{li_tava_2022, weng_humannerf_2022, yu_monohuman_2023} optimize a pose-dependent residual flow field, which acts as a correction to the LBS deformation. SurMo~\cite{hu2024surmo} further improves modeling of clothing by optimizing a surface-based 4D motion representation but requires multi-view videos for supervision. 
However, these methods are optimized per subject, which is computationally expensive and typically requires extensive observations for supervision. Recently developed methods based on 3D Gaussian splatting \cite{qian_3dgs-avatar_2024, pang_ash_2024, kocabas_hugs_2024, hu_gaussianavatar_2024, hu_gauhuman_2024, wen_gomavatar_2024} managed to considerably reduce the optimization time, however, they still do not match the efficiency of feed-forward inference and require more observations for supervision.

\textbf{Generalized neural fields for humans.}
To address the challenges of subject-specific human neural field models, several generalized approaches have been explored \cite{gao_mps-nerf_2024, kwon_neural_2021, hu_sherf_2023, li_ghunerf_2024, masuda_generalizable_2024, chen_geometry-guided_2022, dey_ghnerf_2024, pan_transhuman_2023}. They are particularly efficient at inference time as they only require feed-forward passes through the model. Moreover, they reduce the amount of required observations by leaning on a prior learned during training. They generally condition the neural field on features extracted from the observed views after deforming query points from the target body pose to the observed poses. However, most existing approaches \cite{gao_mps-nerf_2024, kwon_neural_2021, chen_geometry-guided_2022, dey_ghnerf_2024, pan_transhuman_2023} assume that multiple views with the same body pose are observed at test time, which is unlikely to be satisfied in practice. The most related to our work are SHERF \cite{hu_sherf_2023}, GHuNeRF \cite{li_ghunerf_2024}, and GNH \cite{masuda_generalizable_2024}, which, similarly to us, focus on the monocular setting, where each observed view captures a different body pose. SHERF \cite{hu_sherf_2023} reconstructs a human neural field from a single observation by extracting a global feature vector along with local observations aligned using SMPL mesh deformation. GHuNeRF \cite{li_ghunerf_2024} resolves occlusions by projecting the SMPL meshes onto the camera planes. GNH \cite{masuda_generalizable_2024} uses a convolutional renderer instead of volumetric rendering, which increases efficiency but can result in 3D inconsistencies.
The existing generalized approaches to human novel view and novel pose rendering rely predominantly on aligning the observed poses to the canonical pose by deforming the SMPL mesh, where the pose and shape parameters are assumed to be known and accurate. We propose an architecture that does not rely on the SMPL mesh but instead processes the observations densely in 3D and can correct slight pose parameter errors.

%% file: sec/3_method.tex
\section{Method}
\label{sec:method}

Our goal is to learn a generalized dynamic human neural field which can render a novel subject in a given novel pose $\Omega_g$ from given viewpoint $\mathcal{E}_g$ conditioned on $T$ observed views $\mathcal{I} = \{ I_t \}_{t=1}^{T}$. Each view $I_t$ may be capturing the actor in a different pose $\Omega_t$ from an arbitrary viewpoint $\mathcal{E}_t$, where we assume that both $\Omega_t$ and $\mathcal{E}_t$ are known. 
The pose parameters $\Omega_g, \Omega_t$ for the target and observed views contain local joint rotations for $K=24$ joints relative to the \textit{canonical} pose (T-pose).
We also assume that the 3D skeleton shape $J$ in the canonical pose is known, but unlike most previous generalized methods, we do not require the SMPL~\cite{loper_smpl_2015} body shape parameters.
Finally, we assume that camera intrinsics $\mathcal{K}$ and extrinsics $\mathcal{E}_g, \mathcal{E}_t$ are known. 

The overview of our approach is presented in \cref{fig:method-overview}.
We represent the subject's body in a given target pose $\Omega_g$ with a neural radiance field (NeRF)~\cite{mildenhall_nerf_2022} and use standard volumetric rendering to produce the target images from any given viewpoint $\mathcal{E}_g$.
To address the great variety in shapes of posed human bodies, we primarily model the body in the canonical T-pose similar to \cite{weng_humannerf_2022, hu_sherf_2023, yu_monohuman_2023, li_ghunerf_2024} and represent the observation space neural field (corresponding to the target pose) as a deformation of the canonical field.
Specifically, for an observation space query point $x_p \in \mathbb{R}^3$ (corresponding to pose $\Omega$) let $x_c = T_b(x_p, \Omega; \mathcal{I})$ be the corresponding point in the canonical space, where $T_b$ is the \textit{backward deformation} detailed in \cref{sec:method-deformations}.
To query the observation space NeRF for color $\bar{\sigma}$ and density $\bar{\mathbf{c}}$ we instead query the respective canonical fields $\sigma, \mathbf{c}$, \ie $\bar{\sigma}(x_p) = \sigma(x_c)$ and $\bar{\mathbf{c}}(x_p) = \mathbf{c}(x_c)$. Naturally, the canonical and observation-space neural fields are conditioned on the observed views $\mathcal{I}$, which we accomplish by extracting a feature vector $f$ for a canonical query point $x_c$ from $\mathcal{I}$ and setting $(\sigma(x_c; \mathcal{I}), \mathbf{c}(x_c; \mathcal{I})) = \textrm{MLP}(x_c, f)$.

In the remainder of this section, we describe the key contributions in our approach, namely: (1) the conditioning of the canonical neural field on the observed frames (\ie the extraction of features $f$) (\cref{sec:method-conditioning}) and (2) the representation and learning of deformations between the observation and canonical space (\cref{sec:method-deformations}).

\subsection{The Canonical Body Model}
\label{sec:method-conditioning}
To condition the canonical neural field on the observed views $\mathcal{I}$, we extract three types of features: global $f_\textrm{glob}$, voxel-based $f_\textrm{vox}$, and pixel-aligned $f_\textrm{pix}$.
The global and voxel-based features are produced by our 3D \textit{VoluMorph} encoding module, which lifts each observed view of the body (in arbitrary poses) into a partial canonical model and combines these into a single, complete canonical representation at a coarse level.
The pixel-aligned features complement the coarse model with fine details by extracting observations of a query point directly from the posed observed views. 
The three features have complementary strengths and tackle different key challenges: $f_\textrm{vox}$ can resolve occlusions, inject prior and compensate for slight pose inaccuracies; $f_\textrm{glob}$ captures overall characteristics and appearance of the subject through a flat (1D) latent code, which further facilitates prior injection and reconstruction of unobserved regions; $f_\textrm{pix}$, when available, provides direct, high-quality appearance information. Note that the pixel-aligned features are not always relevant (\eg due to occlusions or deformation inaccuracy) or may be unavailable entirely for points that have never been observed. We extract a single global and voxel feature for each point as the encoder already aggregates information from all available observations, but use $T$ pixel-aligned features $f_\textrm{pix}^{(t)}$, one for each observed view. The $f_\textrm{vox}, f_\textrm{glob}$ and $f_\textrm{pix}^{(t)}$ features are combined into a single, final feature vector $f$ for NeRF conditioning by an attention-based feature fusion module.

In this section, let us assume that the backward deformation $T_b$ from the observation space to the canonical space and the forward deformation $T_f$ in the opposite direction are already defined (see \cref{sec:method-deformations} for the definitions). 
We begin the feature extraction process by computing 2D feature maps $F_t = \textrm{CNN}(I_t)$ extracted for each observed view independently with a U-Net feature extractor similar to \cite{yu_monohuman_2023, wang_ibrnet_2021}.

\begin{figure}
    \centering
     \includegraphics[width=\linewidth]{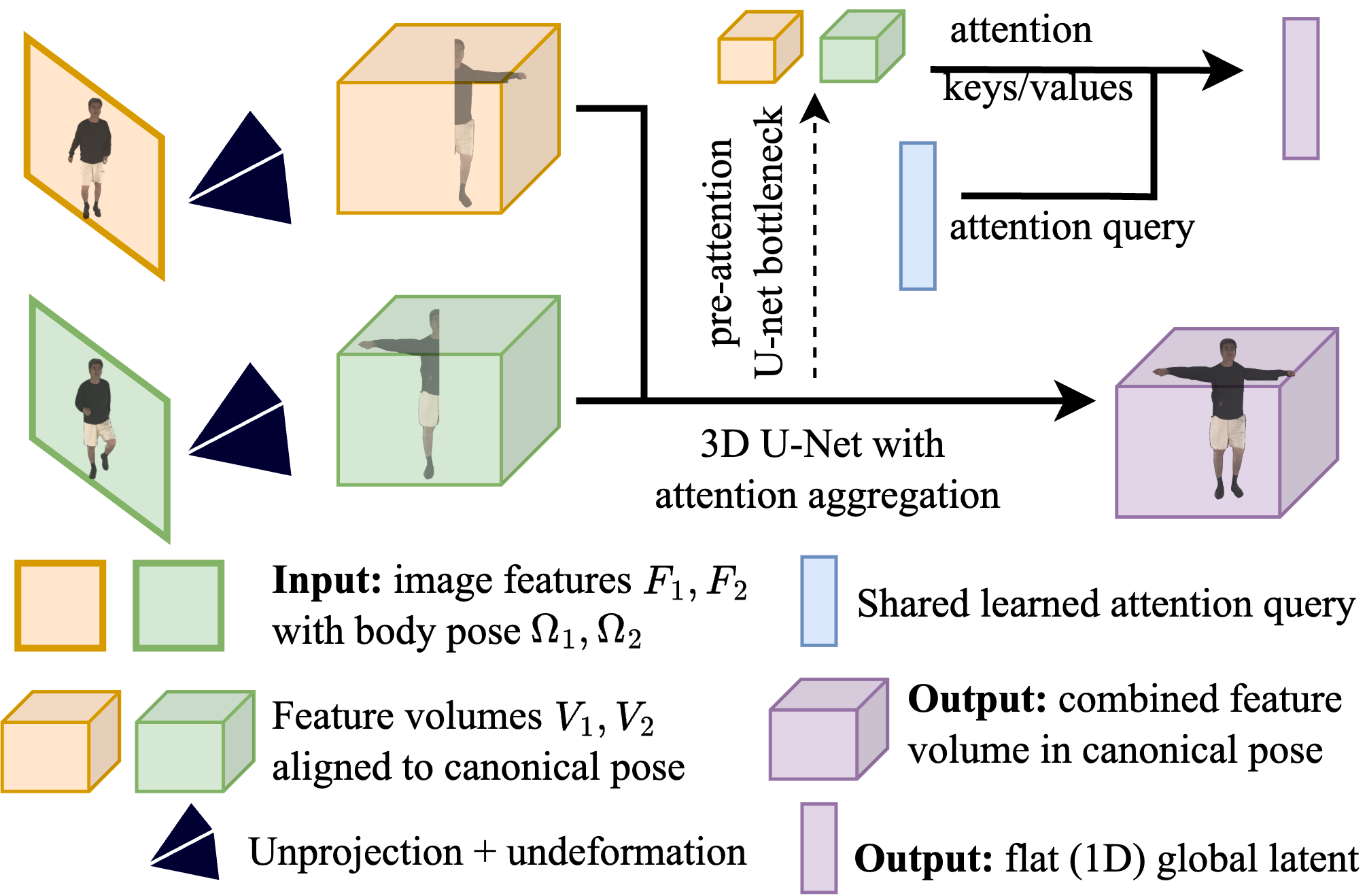}
    \caption{The architecture of our \textit{VoluMorph} module.}
    \label{fig:method-volumorph}
\end{figure}

\textbf{Voxel-based and global features.}
The objective for our 3D \textit{VoluMorph} encoding submodule is to lift the featuremaps $F_t$ corresponding to observed views in arbitrary poses into a combined 3D feature grid, which is aligned with the canonical body pose. 
For an overview of the VoluMorph module see \cref{fig:method-volumorph}.
The initial step in VoluMorph is the \textit{unprojection} of the 2D featuremaps $F_t$ into 3 dimensions based on the known camera parameters.
The initial feature volumes are aligned to the observed poses instead of the canonical pose, which we correct with a volume \textit{undeformation} operation.
Specifically, the combined unprojection and undeformation step is captured as
\begin{equation}
\label{eq:voxel-feature}
    V_t(x_v) = F_t \left[ \pi \left( T_f(x_v, \Omega_t), \mathcal{K}, \mathcal{E}_t \right) \right], 
\end{equation}
where $V_t$ is the undeformed feature grid for view $I_t$ with pose $\Omega_t$, $x_v$ is a 3D grid point, $\pi(\cdot, \mathcal{K}, \mathcal{E}_t)$ is the camera projection operation with instrinsics $\mathcal{K}$ and extrinsics $\mathcal{E}_t$, and $\left[ \cdot \right]$ is bilinear interpolation.
The aligned, partial models (volumes) $V_t$ are combined into a single, complete model $V$ by a 3D U-Net-based convolutional network with attention-based aggregation between views as in \cite{szymanowicz_viewset_2023}. The key feature of this module is that it can learn a semantic understanding of the body and can therefore capture and inject prior knowledge as well as (to some extent) resolve occlusions. We set the voxel feature for a query point $x_c$ in canonical space as $f_\textrm{vox} = V[x_c]$ using trilinear interpolation.

To further facilitate prior injection, we additionally extract a flat 512-dimensional latent code which underlies $f_\textrm{glob}$. To this end, we attend to the volumes of the bottleneck layer of the 3D U-Net. Let $V_t'$ be the volumes corresponding to $V_t$ downscaled to the minimal resolution by the 3D U-Net (before cross-view aggregation). We then get the global latent $z$ by applying a 4-head attention module where the query is a shared, learned 512-dimensional vector and keys/values are linear projections of the flattened $V_t'$ (see \cref{fig:method-volumorph}). We choose to extract the latent $z$ from $V_t'$ as they are already processed for high-level information and abstract away the pose diversity, which simplifies cross-view aggregation. Finally, similar to SHERF \cite{hu_sherf_2023}, we pass the latent $z$ through a style-base encoder \cite{karras_analyzing_2020}, which returns 3 feature planes defining a triplane 3D feature space \cite{chan_efficient_2022}, which is sampled at $x_c$ to get $f_\textrm{glob}$.

\textbf{Pixel-aligned features.} The VoluMorph module operates at a coarse level due to the high memory requirements of voxel-based processing. To effectively increase the rendering resolution, we extract pixel-aligned features $f_\textrm{pix}^{(t)}$ from all observed views for each 3D query point $x_c$ in the canonical space. We utilize the forward deformation $T_f$ to find the corresponding point $x_o^{(t)} = T_f(x_c, \Omega_t)$ in the observation space of view $I_t$. We then extract $f_\textrm{pix}^{(t)}$ for $x_c$ from the featuremap $F_t$ using bilinear interpolation at the projected location $u^{(t)} = \pi(x_o^{(t)}, \mathcal{K}, \mathcal{E}_t)$ (see \cref{fig:method-overview}).

\textbf{Feature fusion.} 
We use an attention-based feature fusion module to determine which (if any) of the pixel-aligned features are relevant and combine them with the coarse model defined by the voxel and global features.
For a query point $x_c$ we extend its $f_\textrm{glob}, f_\textrm{vox}$ and $f_\textrm{pix}^{(t)}$ with spatial information, \ie $x_c$ coordinates, distance from $x_c$ to nearest joint, viewing direction at $x_c$, and, for pixel-aligned features, the viewing direction under which the feature was observed. This information should, intuitively, help the model perform spatial reasoning, \eg rely more on pixel-aligned features when the query and observed viewing directions align.
The extended features are processed with a single transformer encoder layer.
The final feature $f$ for $x_c$ is constructed by selecting the most relevant features using an attention layer, where queries are based on the spatial features of $x_c$ (position, viewing direction, etc.) and keys/values are based on the transformer encoder's output.

\subsection{Deformation representation and learning}
\label{sec:method-deformations}
Recall that our definition of the neural field in observation space relies on the \textit{backward deformation} $T_b$, which for a point $x_p$ in observation space with pose $\Omega$ returns a corresponding point $x_c = T_b(x_p, \Omega ; \mathcal{I})$ in the canonical space.
Moreover, our feature extraction process (\cref{sec:method-conditioning}) relies on the \textit{forward deformation} $T_f$, which for a point in the canonical space returns a corresponding point in the observation space (opposite to $T_b$). In this section, we explain how these deformations are parameterized and conditioned on the observed views.

\textbf{Linear Blend Skinning for Humans.}
Our representation follows \cite{weng_vid2actor_2020, weng_humannerf_2022, yu_monohuman_2023} but is adapted to the generalized setting with a feed-forward conditioning mechanism. The deformations follow the linear blend skinning (LBS) model
\begin{align}
    \label{eq:tf} T_b(x_p, \Omega ; \mathcal{I}) &= \sum_{i=1}^K w_p^{(i)}(x_p ; \mathcal{I}) (R_i x_p + t_i), \\
    \label{eq:tb} T_f(x_c, \Omega ; \mathcal{I}) &= \sum_{i=1}^K w_c^{(i)}(x_c ; \mathcal{I}) R_i^{-1} (x_c - t_i),
\end{align}
where $ w_p^{(i)}, w_c^{(i)}$ are the blend (motion) weights for the $i$-th bone in the posed and canonical space (respectively), and $R_i, t_i$ are the rotation and translation which transform points on the $i$-th bone from observation space to canonical. The rotation $R_i$ and translation $t_i$ are explicitly computed from the pose $\Omega$ (see \cite{weng_vid2actor_2020, weng_humannerf_2022} for a derivation). Following \cite{weng_humannerf_2022}, we only optimize the canonical motion weights $w_c^{(i)}$ and express motion weights in the posed space as
\begin{equation}
\label{eq:wp}
    w_p^{(i)}(x_p; \mathcal{I}) = \frac{w_c^{(i)}(R_i x_p + t_i; \mathcal{I})}{\sum_{k=1}^K w_c^{(k)}(R_k x_p + t_k; \mathcal{I})}
\end{equation}
to improve generalization across poses. Throughout our pipeline, the motion weights are represented as a discrete volume $W(\mathcal{I})$ with $K$ channels summing to one at each voxel, where $K$ is the number of joints in $J$.
The continuous motion weights $w_c$ are then computed using trilinear interpolation from the discrete volume.

\textbf{Conditioning deformations on observed views.} The deformations $T_b, T_f$ are conditioned on the observed views $\mathcal{I}$ through motion weights $w_c$. Following \cite{weng_vid2actor_2020, weng_humannerf_2022}, we begin with a heuristic initial estimate of the motion weights $W_0$ constructed by placing ellipsoidal Gaussians around the $i$-th bone for the $i$-th channel of $W_0$. We additionally refine the initial weights with a learned bias represented as a 3D CNN output from a random fixed latent code (shared between all subjects).
The initial motion weights define through \cref{eq:tf} an initial guess $\tilde{T}_f$ for the forward deformation.  We then estimate an observation-conditioned correction $\Delta W(\mathcal{I})$ to the initial $W_0$ using our VoluMorph module introduced in \cref{sec:method-conditioning} (without the global latent) with $\tilde{T}_f$ as the forward deformation (in \cref{eq:voxel-feature}).
The intuition behind this choice is that $\tilde{T}_f$ will provide a rough alignment of observed views to the canonical pose from which the convolutional architecture of VoluMorph will recover the body shape (and through that the motion weights) despite some remaining misalignment.
Note that, as shown in \cref{fig:method-overview}, we use two separate VoluMorph modules, one for the neural field features $V$ and one for the motion weights $\Delta W$. 

\subsection{Network training}
We optimize all components of our model end-to-end to minimize the loss function
\begin{equation}
    \mathcal{L} = \mathcal{L}_\textrm{LPIPS} + \lambda_1 \mathcal{L}_\textrm{MSE} + \lambda_2 \mathcal{L}_\textrm{consis} + \lambda_3 \mathcal{L}_\textrm{near}
\end{equation}
where $\mathcal{L}_\textrm{MSE}$ is the pixel-wise mean squared error from the ground-truth image and $\mathcal{L}_\textrm{LPIPS}$ is the LPIPS \cite{zhang_unreasonable_2018} perceptual loss. We include the $\mathcal{L}_\textrm{consis}$ term proposed by MonoHuman~\cite{yu_monohuman_2023}, which regularizes the motion weights by  
minimizing $\lVert x_p - T_f(T_b(x_p, \Omega), \Omega) \rVert^2$. 

Furthermore, we found that the feed-forward prediction of motion weights can sometimes result in unnatural motions, especially when the pose parameters are noisily estimated from the observed images.
To mitigate this, we introduce additional regularization guidance
\begin{equation}
\label{eq:reg-loss}
    \mathcal{L}_\textrm{near} = \sum_{x \in W} \sum_{k=1}^K w_c^{(k)} d(x, B_k),
\end{equation}
where $x \in W$ are voxel positions in the motion weights volume $W$ and $d(x, B_k)$ is the distance from $x$ to the $k$-th bone (line segment), which encourages assigning points to their nearest bone.
We use $\lambda_1 = 0.3, \lambda_2 = 2, \lambda_3 = 0.1$ in our experiments.

%% file: sec/4_experiments.tex
\begin{table*}
    \centering
    \begin{tabular}{l| c c | c c | c c | c c}
        \hline
        \multirow{3}{*}{Method} & \multicolumn{4}{c|}{HuMMan \cite{cai_humman_2022}} & \multicolumn{4}{c}{DNA-Rendering \cite{cheng_dna-rendering_2023}} \\
        \cline{2-9}
        & \multicolumn{2}{c|}{\textit{Accurate body params.}} & \multicolumn{2}{c|}{\textit{Estim. body params.}} & \multicolumn{2}{c|}{\textit{Accurate body params.}} & \multicolumn{2}{c}{\textit{Estim. body params.}} \\
        \cline{2-9}
        & PSNR $\uparrow$ & LPIPS* $\downarrow$ & PSNR $\uparrow$ & LPIPS* $\downarrow$ & PSNR $\uparrow$ & LPIPS* $\downarrow$ & PSNR $\uparrow$ & LPIPS* $\downarrow$ \\
        \hline
        SHERF (Mo) & 26.95 & 44.12 & 24.23 & 61.44 & 28.49 & 48.22 & 26.93 & 61.97 \\
        \hline
        GHuNeRF+ (1 obs.) & 23.89 & 44.00 & 23.17 & 50.24 & 26.59 & 53.10 & 26.19 & 56.46 \\
        GHuNeRF+ (2 obs.) & 23.97 & 43.72 & 23.27 & 49.96 & 26.69 & 52.92 & 26.28 & 56.16 \\
        GHuNeRF+ (4 obs.) & 24.00 & 43.66 & 23.31 & 49.86 & 26.70 & 52.93 & 26.31 & 56.11 \\
        GHuNeRF (4 obs.) & 23.89 & 63.02 & 23.36 & 68.76 & 27.78 & 69.71 & 27.28 & 74.54 \\
        \hline
        \textbf{Ours} (1 observed) & 26.70 & 33.43 & 25.08 & 42.28 & 27.86 & 40.25 & 27.00 & 47.21 \\
        \textbf{Ours} (2 observed) & 27.38 & 30.20 & 25.33 & 40.93 & 28.35 & 38.03 & 27.31 & 45.45 \\
        \textbf{Ours} (3 observed) & 27.64 & 28.88 & \textbf{25.40} & 40.53 & 28.63 & \textbf{36.88} & 27.45 & 44.79 \\
        \textbf{Ours} (4 observed) & \textbf{27.66} & \textbf{28.72} & \textbf{25.40} & \textbf{40.52} & \textbf{28.65} & 36.89 & \textbf{27.46} & \textbf{44.76} \\
        \hline
    \end{tabular}
    \caption{Quantitative comparison of our method with SHERF \cite{hu_sherf_2023} and GHuNeRF \cite{li_ghunerf_2024} with various numbers of observed views. SHERF (Mo) is trained in our monocular framework, and GHuNeRF+ contains the added LPIPS loss (see \cref{sec:evaluation-protocol}). $\textrm{LPIPS*} = \textrm{LPIPS} \times 10^3$.}
    \label{tab:quantitative-comparison-base}
\end{table*}

\begin{figure*}
    \centering
    \includegraphics[width=0.98\linewidth]{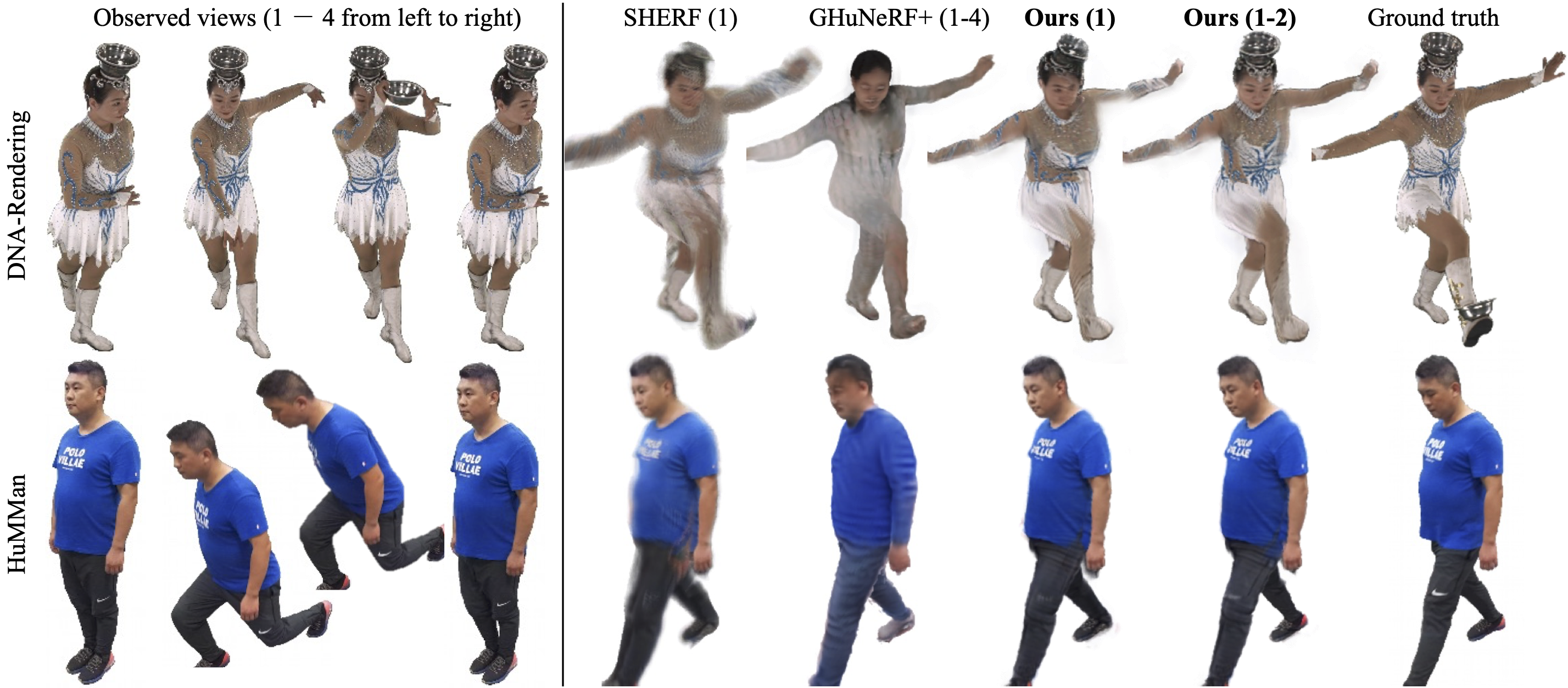}
    \caption{Qualitative comparison with SHERF (Mo) \cite{hu_sherf_2023} and GHuNeRF+ \cite{li_ghunerf_2024} given accurate shape and pose parameters. Numbers in parentheses indicate the range of observed views supplied to the respective models.}
    \label{fig:qualitative-results}
\end{figure*}

\begin{figure*}
    \centering
    \includegraphics[width=\linewidth]{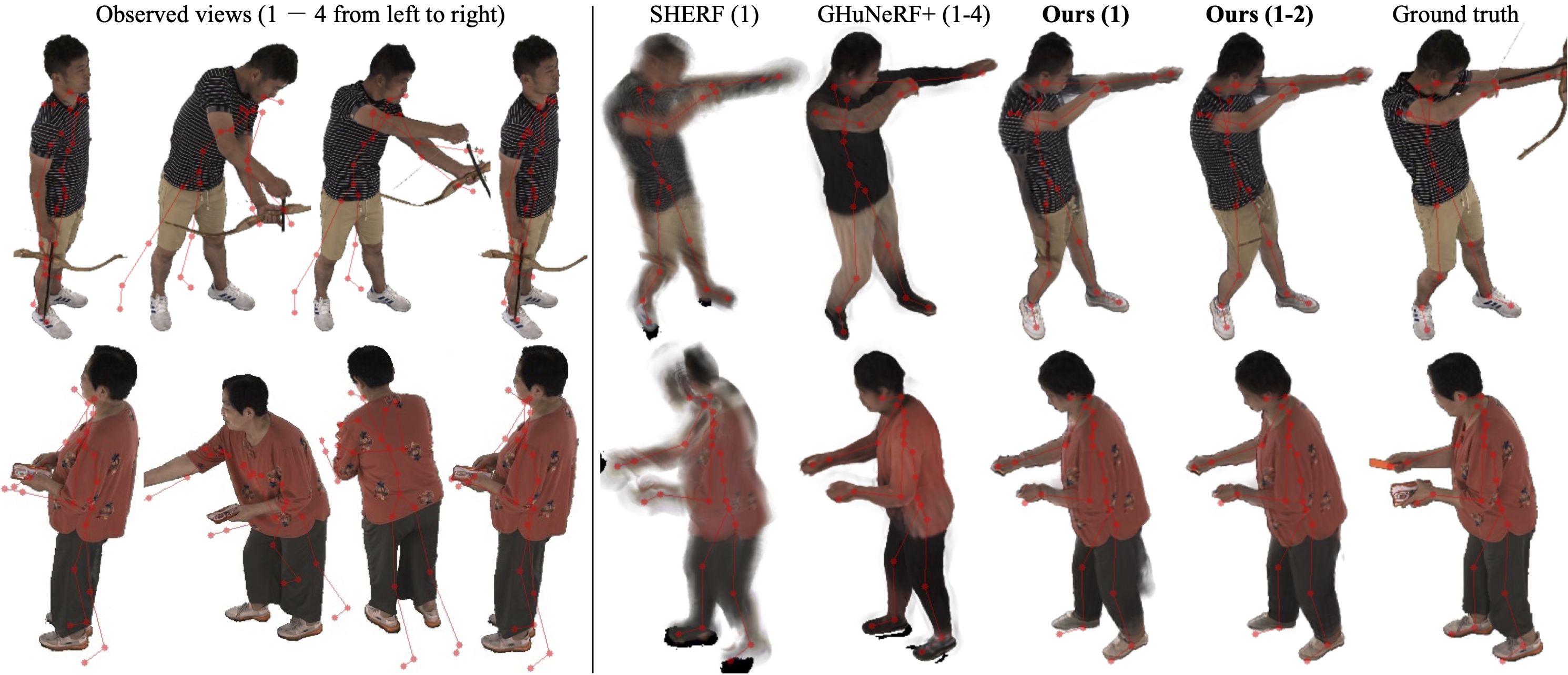}
    \caption{Qualitative comparison with SHERF (Mo) \cite{hu_sherf_2023} and GHuNeRF+ \cite{li_ghunerf_2024} on DNA-Rendering \cite{cheng_dna-rendering_2023} with shape and pose parameters estimated from observed views. Numbers in parentheses indicate the range of observed views supplied to the respective models.}
    \label{fig:qualitative-eval-estim}
\end{figure*}

\section{Experiments}
\label{sec:experiments}

We evaluate our approach on two large-scale human motion datasets: HuMMan \cite{cai_humman_2022} and DNA-Rendering \cite{cheng_dna-rendering_2023}. The HuMMan dataset consists of 339 motion sequences performed by 153 subjects. We use 317 sequences for training and 22 sequences for hold-out evaluation according to the official data split. We use parts 1 and 2 from the DNA-Rendering dataset, which together consist of 436 sequences performed by 136 subjects.
We split the sequences into 372 for the training set and 64 for the test set. We assign all sequences performed by the same actor to either the train or test set, which results in 118 training subjects and 18 test subjects. 
To reduce the computational requirements, we subsample the camera sets from 10 to 8 on HuMMan and from 48 to 6 on DNA-Rendering (see \cref{sec:suppl-experiments} details).

\subsection{Implementation Details}
To simulate practical conditions where multi-view synchronized videos are unlikely to be available, we train our models in the fully monocular framework. Specifically, during training, the observed frames are selected from the same camera as the target frame.
At training time, we provide our model with 2 randomly sampled observed frames at each step and render 32x32 patches of the target frames downscaled to 0.25 of the original resolution.

\textbf{Architecture.} We use the same U-Net feature extractor as \cite{yu_monohuman_2023}.
We use a voxel grid with a resolution of 32 in our VoluMorph modules. The architecture of the 3D U-Net convolutional network inside our VoluMorph modules follows \cite{szymanowicz_viewset_2023}.
We adopt the style-based encoder from~\cite{hu_sherf_2023} to decode the global feature triplanes.
The 2D featuremaps $F_t$ and 3D features $f_\textrm{glob}, f_\textrm{vox}, f_\textrm{pix}$ are 32-dimensional.
All attention layers in the feature fusion module use 4 heads and have internal dimension 64.
We use an 8-layer MLP NeRF decoder with 256-dimensional hidden layers.
See \cref{sec:suppl-arch} for more details on architecture and optimization.

{
\setlength{\tabcolsep}{5pt}
\begin{table*}
    \centering
    \begin{tabular}{l| c c | c c | c c | c c}
        \hline
        \multirow{3}{*}{Method} & \multicolumn{4}{c|}{\textit{Accurate body parameters}} & \multicolumn{4}{c}{\textit{Estimated body parameters}} \\
        \cline{2-9}
        & \multicolumn{2}{c|}{1 observed} & \multicolumn{2}{c|}{2 observed} & \multicolumn{2}{c|}{1 observed} & \multicolumn{2}{c}{2 observed} \\
        \cline{2-9}
        & PSNR $\uparrow$ & LPIPS* $\downarrow$ & PSNR $\uparrow$ & LPIPS* $\downarrow$ & PSNR $\uparrow$ & LPIPS* $\downarrow$ & PSNR $\uparrow$ & LPIPS* $\downarrow$ \\
        \hline
        Baseline & 27.52 & 53.16 & 27.92 & 49.47 & 22.21 & 61.42 & 22.35 & 59.52 \\
        + $\Delta W$ & 27.58 & 43.61 & 28.16 & 40.46 & 26.45 & 51.62 & 26.85 & 49.50 \\
        + $f_\textrm{vox}$ & 27.69 & 40.73 & 28.15 & 38.49 & 26.21 & 54.72 & 26.50 & 51.59 \\
        + $\Delta W + f_\textrm{vox}$ & 27.83 & 40.27 & 28.31 & \textbf{37.80} & 26.48 & 51.06 & 26.87 & 48.26 \\
        + $\Delta W + f_\textrm{vox} + f_\textrm{glob}$ & 27.82 & 40.37 & 28.30 & 38.13 & 26.62 & 48.83 & 27.00 & 47.01 \\
        + $\Delta W + f_\textrm{vox} + f_\textrm{glob} + \mathcal{L}_\textrm{near}$& \textbf{27.86} & \textbf{40.25} & \textbf{28.35} & 38.03 & \textbf{27.00} & \textbf{47.21} & \textbf{27.31} & \textbf{45.45} \\
        \hline
    \end{tabular}
    \caption{Ablation study results on DNA-Rendering \cite{cheng_dna-rendering_2023} using accurate and estimated body parameters. See \cref{sec:exp-ablation} for a description of the components. $\textrm{LPIPS*} = \textrm{LPIPS} \times 10^3$.}
    \label{tab:ablation}
\end{table*}
}

\subsection{Evaluation Protocol}
\label{sec:evaluation-protocol}
We compare our approach to two state-of-the-art methods: SHERF \cite{hu_sherf_2023} and GHuNeRF \cite{li_ghunerf_2024}.
In the evaluations, we follow our monocular protocol, \ie we provide observed frames from the same camera as the target frame. 
Hu et al. train SHERF \cite{hu_sherf_2023} in a multi-view framework, hence, to ensure a fair comparison, we re-train SHERF in our monocular framework and we refer to it as SHERF (Mo).
We also re-train the GHuNeRF model since it was not originally evaluated on the datasets used in this work. Moreover, the original GHuNeRF 
does not include a perceptual LPIPS~\cite{zhang_unreasonable_2018} loss term, and for a fair comparison, we also report results for GHuNeRF+, which is our version of GHuNeRF with the perceptual LPIPS term added to the loss.
For each baseline training, we match the hyperparameters as closely as possible to the ones used by the respective authors. We train GHuNeRF using 4 observed views as it was originally designed to work with many input frames.

To measure the rendering quality given \textit{novel poses}, we split the motion sequences approx. in half at frame $\lfloor \frac{T+1}{2} \rfloor$, where $T$ is the sequence length, and provide observations from the first half, while we render frames from the second half (see \cref{sec:suppl-experiments} for details).
Note that even though the datasets used in this work use stationary cameras, \textit{HumMorph} admits a different viewpoint for each frame (observed and target). Moreover, the evaluation datasets contain sequences where the global body orientation changes with respect to the camera, which is equivalent to changing the viewpoint.
We evaluate the rendering quality using two common metrics: pixel-wise PSNR and the perceptual LPIPS metric.

In our experiments, we consider two scenarios depending on whether accurate body parameters are available for the observed frames. If they are not, we use an off-the-shelf HybrIK \cite{li_hybrik_2021} model to estimate the SMPL \cite{loper_smpl_2015} shape and pose parameters for each observed frame independently. For this experiment, we re-train our models and the baselines using a mixture of accurate and estimated parameters (see \cref{sec:suppl-estim} for details).
At test time, we provide the models with estimated parameters for the observed frames but use accurate poses for the target frame (the target pose \textit{cannot} be estimated). The shape parameters for the target frame are not provided and must be estimated from the observed views.
We use the ground-truth camera poses and leave integration of camera pose estimation as future work.

\subsection{Quantitative comparison}
The results of our quantitative evaluation are presented in \cref{tab:quantitative-comparison-base}. Our method achieves a significantly better perceptual score (LPIPS) compared to SHERF and GHuNeRF on both datasets given even a single observed view. However, with 1 observation, our PSNR score is below that of SHERF, which we attribute to the fact that PSNR favours oversmoothed results over slight misalignments. Many subjects in the datasets are dressed in complex clothing with intricate details and failing to properly model its folding dynamic, which none of the methods are targeting, can significantly lower the PSNR score. 
Our method outperforms SHERF in both PSNR and especially LPIPS given 2 views on HuMMan and 3 views on DNA-Rendering. Note that SHERF can only be conditioned on a single view.
Moreover, \cref{tab:quantitative-comparison-base} shows that, while the quality of all models degrades when using estimated parameters, our method achieves a better perceptual score compared to the baselines provided with \textit{accurate} parameters. 
Finally, the performance of our method on both metrics generally improves as it is given additional observations, but it saturates at 3 views. Note that in our experiments, additional views beyond the 3rd often do not considerably increase pose diversity. 
While the performance of GHuNeRF+ also increases given additional views, this effect is less pronounced.

\subsection{Qualitative results}
\label{sec:quali-results}
\cref{fig:qualitative-results} and \cref{fig:qualitative-eval-estim} show qualitative comparisons with the baselines given accurate and estimated body parameters (respectively), with varying numbers of observed frames.
As seen in \cref{fig:qualitative-results}, our renders are overall more sharp and detailed compared to both SHERF and GHuNeRF+. Given the single observation, SHERF struggles to resolve occlusions or impose a smoothness prior, which results in `phantom' limbs imprinted on the torso. Although in some cases our method exhibits a similar issue, it can combine two observations to better match the body geometry and remove the artifacts. In general, GHuNeRF+ produces oversmoothed results, which do not reproduce details and only loosely reconstruct the original appearance.
As shown in \cref{fig:qualitative-eval-estim}, our method successfully reconstructs clothing details even when provided with estimated body poses. Moreover, despite considerable pose estimation errors, it can combine the two input views to refine the body model (see top row \cref{fig:qualitative-eval-estim}). See \cref{sec:add-results} for more qualitative results.

\subsection{Ablation study}
\label{sec:exp-ablation}
\cref{tab:ablation} presents results of an ablation study, which validates the design and usage of the main components of our model. The baseline model does not include the motion weights correction module $\Delta W$ and relies only on the pixel-aligned features. We then subsequently add further components, namely motion weights correction $\Delta W$, the voxel-features $f_\textrm{vox}$, the global feature $f_\textrm{glob}$ and the $\mathcal{L}_\textrm{near}$ regularization (see \cref{eq:reg-loss}).
As shown in \cref{tab:ablation}, including the voxel features module grants the largest boost in rendering quality with accurate body parameters. Without them, the model will struggle to resolve occlusions and inject prior, which can result in rendering artifacts. Combining that with the motion weights correction improves results further, although by a smaller margin. Note that with accurate body parameters, use of the global feature is not required, however, we include it in our final model for consistency.
In contrast, both the motion weights correction and the global feature have a significant impact when using estimated body parameters. Finally, including the $\mathcal{L}_\textrm{near}$ regularization refines the results further due to encouraging natural deformations.
Additionally, \cref{tab:ablation} relates the positive impact of our design choices to that of supplying the second observation.

%% file: sec/5_conclusion.tex
\section{Conclusion}
\label{sec:conclusion}
We introduced \textit{HumMorph}, a novel generalized approach for free-viewpoint synthesis of human bodies with explicit pose control conditioned on a variable number of views. We demonstrated results of state-of-the-art perceptual visual quality given a single observed view and a significant boost in quality when two conditioning views are available. 
We also demonstrated that our approach is significantly more robust to inaccuracies in noisily estimated body pose parameters compared to prior methods.

\textbf{Limitations and future work.} 
Despite the increased robustness of our approach to pose estimation errors, the resulting renders still show considerable room for improvement.
Moreover, in this work we use ground truth camera poses. Investigating the use of estimated camera poses and adjusting the model accordingly is an interesting direction for future work. 
HumMorph also does not explicitly model deformations of clothing and usually cannot reconstruct interactions with objects, which requires further research.

\noindent \textbf{Acknowledgments:} This work was supported by the United Kingdom Research and Innovation (grant EP/S023208/1), UKRI Centre for Doctoral Training in Robotics and Autonomous Systems at the University of Edinburgh, School of Informatics. HB was supported by the EPSRC Visual AI grant EP/T028572/1.

%% file: sec/X_suppl.tex
\appendix
\maketitlesupplementary
\renewcommand{\thesection}{\Alph{section}}

\section{Additional Details Regarding the Method}
\textbf{Rendering equations.}
In \cref{sec:method} we define the density and color functions $\bar{\sigma}, \bar{\mathbf{c}}$ of a NeRF in observation space corresponding to pose $\Omega$.
We use volumetric rendering to synthesize the target image. Specifically, the color of each pixel $u$ in the rendering is computed as follows:
\begin{align}
    \label{eq:nerf-rend} C(u) &= \sum_{i=1}^M \left[ \prod_{j=1}^{i-1} \left( 1-\alpha_j \right) \right] \alpha_i \bar{\mathbf{c}}(x_i, \Omega), \\
    \label{eq:nerf-alpha} \alpha_i &= \left(1 - \exp \left( \bar{\sigma}(x_i, \Omega) \Delta x_i \right) \right),
\end{align}
where $x_i \in \mathbb{R}^3$ for $1 \leq i \leq M$ are points along ray $r_u$ passing through pixel $u$ in the image plane and $\Delta x_i = \lVert x_{i+1} - x_i \rVert_2$. Following HumanNeRF \cite{weng_humannerf_2022}, we only sample the query points $x_i$ inside a 3D bounding box estimated from the human skeleton in pose $\Omega$.

\begin{figure}
    \centering
    \includegraphics[width=\linewidth]{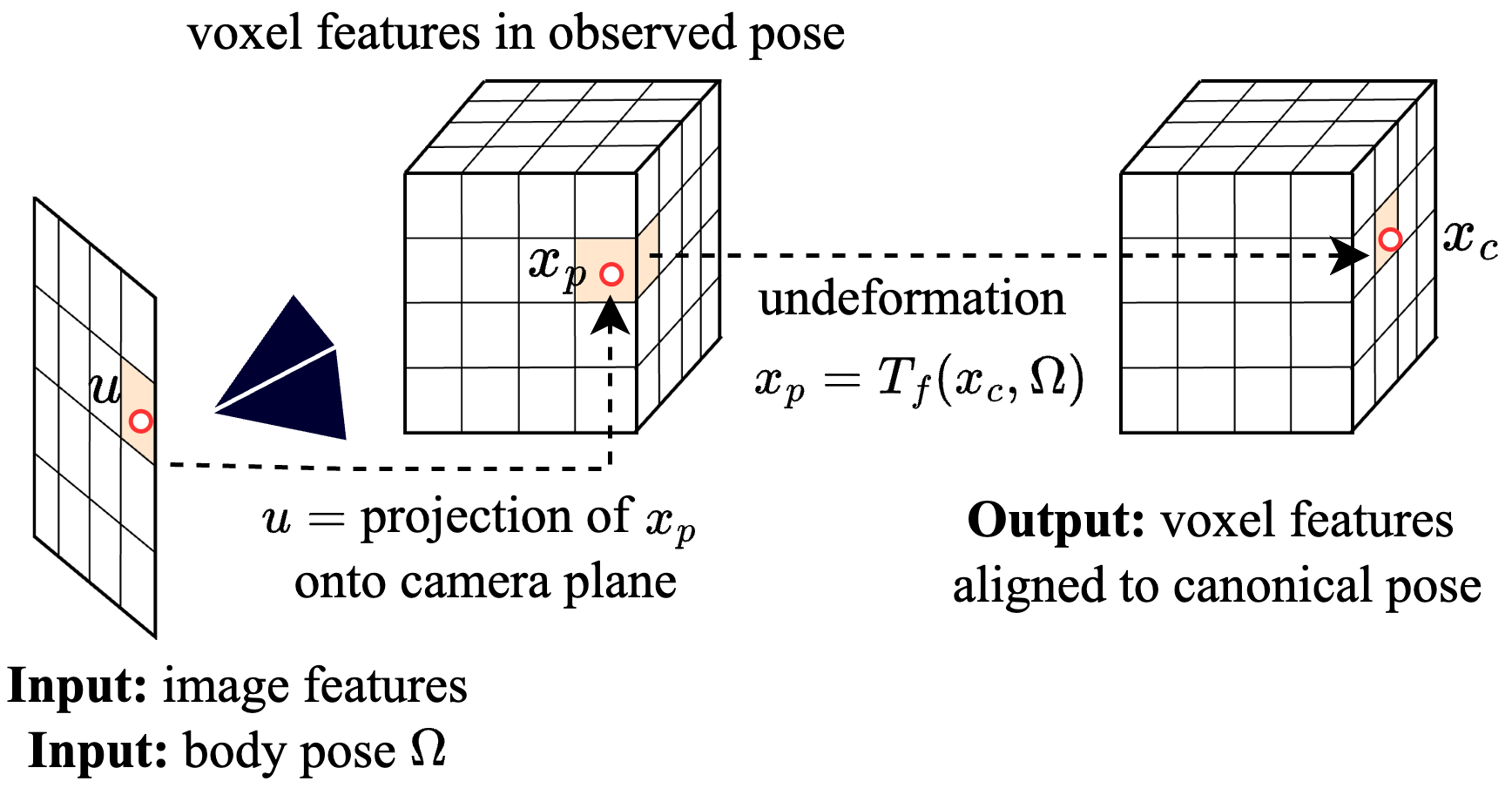}
    \caption{Diagram of the unprojection and undeformation operation defined by \cref{eq:voxel-feature}.}
    \label{fig:unnproj-undeform}
\end{figure}

\textbf{Unprojection and undeformation.} See \cref{fig:unnproj-undeform} for an illustration of the unprojection and undeformation operation defined by \cref{eq:voxel-feature}.

\textbf{Network training and loss functions.}
In a single training step, we render $G=6$ patches $P_i$ of size $H \times H$ with $H = 32$, which are used to compute $\mathcal{L}_\textrm{LPIPS}$ with a VGG backbone. We also have
\begin{equation}
    \mathcal{L}_\textrm{MSE} = \frac{1}{G \cdot H^2} \sum_{i=1}^G \sum_{u \in P_i} \lVert C(u) - \hat{C}(u) \rVert_2^2,
\end{equation}
where $u$ is a pixel in patch $P_i$, $C(u)$ is the rendered color of $u$ (as in \cref{eq:nerf-rend}) and $\hat{C}(u)$ is the ground truth color of $u$.
The deformation consistency component $\mathcal{L}_\textrm{consis}$ encourages consistency between the forward and backward deformations $T_f, T_b$ (respectively; see \cref{sec:method-deformations}). Recall that, intuitively, we should have $T_f(T_b(x_c, \Omega), \Omega) = x_c$ for a point $x_c$ in canonical space and pose $\Omega$. However, with the LBS deformation model, this condition is rarely satisfied and it depends on the motion weights $W$. Following MonoHuman~\cite{yu_monohuman_2023}, we include
\begin{equation}
    \mathcal{L}_\textrm{consis} = 
    \begin{cases}
        d \quad \textrm{if} \quad d \geq \eta, \\
        0 \quad \textrm{otherwise},
    \end{cases}
\end{equation}
where $d =\lVert x_p - T_f(T_b(x_p, \Omega), \Omega) \rVert_2^2$ for a point $x_p$ in the observation space with pose $\Omega$, in the loss function to regularize the motion weights. We compute $\mathcal{L}_\textrm{consis}$ on all query points used in volumetric rendering and use $\eta = 0.05$.

\section{Additional Implementation Details}
\label{sec:suppl-arch}
To better preserve the low-level information, we concatenate the feature maps $F_t$ with resized input images $I_t$. Hence, the pixel-aligned features $f_\textrm{pix}$ have dimensionality $32 + 3$.
Both the motion weights and the voxel features VoluMorph submodules output a 32-dimensional voxel grid of size 32 along the $X, Y$ axes and size 16 along the $Z$ axis, which corresponds to the human body shape.
The output of the voxel features VoluMorph submodule is directly sampled to create $f_\textrm{vox}$ features, which are also 32-dimensional.
The output feature volume for motion weights correction is additionally projected (coordinate-wise) into $K = 24$ channels (one per joint) using a $1 \times 1$ convolution. The output of the convolution is the observation-conditioned correction $\Delta W(\mathcal{I})$ in log-space, which is combined with the initial estimate $W_0$ as follows
\begin{equation}
W(\mathcal{I}) = \textrm{softmax}(\Delta W(\mathcal{I}) + \log W_0).   
\end{equation}

\textbf{Feature fusion module.}
Here we provide additional details on the implementation of the feature fusion module introduced in \cref{sec:method-conditioning}. 
Let $x_c$ be a query point in canonical space. 
We describe how our feature fusion module computes the final feature $f$ for a single $x_c$, which, in practice, is applied independently to each query point.

The feature vectors are first extended with positional encodings of spatial information on the query point $x_c$: its coordinates, the viewing direction on $x_c$ in the target render transformed to canonical space, and the vector from $x_c$ to the nearest joint in the skeleton. We additionally append the motion weights $W$ sampled at $x_c$, which serve as proxy information on the body shape. For the pixel-aligned features, we also append the viewing direction (transformed to canonical space) under which the features were observed.
The extended features are then aligned using two separate 2-layer MLPs with hidden dimensions 128 and output dimensions 64. The aligned features are processed by a transformer encoder layer with 4 attention heads and internal dimension 64. 

The standalone spatial information on the query point (\ie coordinates, viewing direction, vector to the nearest joint, and sampled motion weights) is aligned with the features using a 2-layer MLP with hidden dimension 128 and output dimension 64. The final feature $f$ is computed with a 4-head attention layer with internal dimension 64, where the (aligned) standalone spatial information on $x_c$ is used as a query and the transformer encoder's outputs are used as keys/values.

\textbf{Optimization.} We optimize the parameters of our model using the Adam optimizer with learning rate $2 \times 10^{-5}$ for the motion weights correction submodule and $2 \times 10^{-4}$ for the rest. We additionally delay the optimization of the motion weights module until iteration 5K.
We found that optimizing the motion weights end-to-end with the rest of the pipeline can, in some cases, introduce training instabilities, which we contain by clipping the loss gradients to L2 norm of 7.5.
We run our optimization for $\sim$300K iterations on 4 NVIDIA RTX 6000 GPUs, which takes about 5 days.

\section{More Details on the Experiments}
\label{sec:suppl-experiments}
\textbf{Selection of cameras.} To reduce the computational cost of running our experiments, we subsample the camera sets of both datasets. For training and evaluation on the HuMMan dataset~\cite{cai_humman_2022} we drop the cameras with indices 2 and 7 (the ones with the highest vertical position). 
For training on the DNA-Rendering dataset~\cite{cheng_dna-rendering_2023} we keep cameras with index $c$ such that $c  \equiv 1 \mod 4$ (12 cameras total), while for evaluation we use cameras with index $c$ such that $c  \equiv 1 \mod 8$ (6 cameras total).
We use the same camera subset for training and evaluation of all models, including baselines.

\textbf{Image resolution.} 
During training of our method on both datasets, we render the frames (patches) at $\frac{1}{4}$ of the original resolution, \ie $480 \times 270$ for the HuMMan dataset and $512 \times 612$ for the DNA-Rendering dataset. We train SHERF~\cite{hu_sherf_2023} and GHuNeRF \cite{li_ghunerf_2024} on the HuMMan dataset using $\frac{1}{3}$ of the original resolution, \ie $634 \times 356$ and using $\frac{1}{4}$ of the original resolution on the DNA-Rendering dataset, \ie $512 \times 612$.
We evaluate our method and the baselines using $\frac{1}{3}$ of the original resolution on the HuMMan dataset and using $\frac{1}{4}$ of the original resolution on the DNA-Rendering dataset.

\textbf{Subsampling frames.}
We subsample the frames of all motion sequences in the DNA-Rendering dataset~\cite{cheng_dna-rendering_2023} to a maximum of 30 frames per sequence. We perform the subsampling at constant intervals across the full length of each sequence.
We use the full sequences in the HuMMan dataset~\cite{cheng_dna-rendering_2023}.

\textbf{Selection of observed frames.}
During training, our models are provided with $T=2$ observed frames, which are uniformly sampled from the full motion sequence (without the target frame). The observed frames are sampled from the same camera as the target frame. During monocular training, SHERF~\cite{hu_sherf_2023} (Mo) is provided with a random frame (except the target frame) from the same camera as the target frame. 
GHuNeRF during training is supplied with 4 randomly sampled observed frames.

For evaluation, we split the motion sequences approximately in half at frame $\lfloor \frac{T+1}{2} \rfloor$, where $T$ is the sequence length, and provide observations from the first half, while we measure the quality of reconstruction on the frames from the second half. Specifically, when $T$ is the motion sequence length (in frames), the observed frames are selected based on the table below:

\begin{center}
\begin{tabular}{c|c }
     Num. observ. & Indices of observed frames \\
     \hline
     1 & 0 \\
     2 & 0, $\lfloor T \cdot \frac{1}{4} \rfloor$ \\
     3 & 0, $\lfloor T \cdot \frac{1}{4} \rfloor$, $\lfloor T \cdot \frac{3}{8} \rfloor$ \\
     4 & 0, $\lfloor T \cdot \frac{1}{4} \rfloor$, $\lfloor T \cdot \frac{3}{8} \rfloor$, $\lfloor T \cdot \frac{1}{8} \rfloor$ \\
\end{tabular}
\end{center}
Note that, as SHERF~\cite{hu_sherf_2023} only accepts a single observed frame, in the quantitative experiments it is provided with the first frame of each sequence. We provide qualitative results of SHERF given other observed frames. In the qualitative results, the index of the observed frame number $i$ is the last entry of row $i$ in the table above.

\subsection{Estimated Body Shape and Pose Parameters}
\label{sec:suppl-estim}
To obtain the estimated SMPL~\cite{loper_smpl_2015} pose and shape parameters, we use an off-the-shelf HybrIK~\cite{li_hybrik_2021} model for each frame in the motion sequences independently.
We then re-train our models, SHERF (Mo) and GHuNeRF(+) using a mixture of accurate and estimated parameters.
At each training step, we use the estimated parameters with probability $p$ or the accurate parameters with probability $1-p$, where $p$ increases linearly throughout the training from 0 at the beginning to 0.75 at roughly half of the training process.

When using estimated body parameters, during both training and evaluation, we provide the models with the estimated body shape and pose parameters for the observed frames. However, we always provide accurate pose parameters for the target frames, which is motivated by the practical scenario, where pose parameters are either transferred from a different motion or generated with a separate model. Furthermore, since the target frame is not known in practice, estimating the target pose is not meaningful. In contrast, the body shape is always assumed to be unknown and, therefore, has to be estimated from the observed frames.
Note that, in this experiment, we use the ground-truth camera poses for both models.

\begin{table*}
    \centering
    \begin{tabular}{l| c c c | c c c}
        \hline
        \multirow{2}{*}{Method} & \multicolumn{3}{c|}{\textit{Accurate body parameters}} & \multicolumn{3}{c}{\textit{Estimated body parameters}} \\
        \cline{2-7}
        & PSNR $\uparrow$ & LPIPS* $\downarrow$ & SSIM $\uparrow$ & PSNR $\uparrow$ & LPIPS* $\downarrow$ & SSIM $\uparrow$ \\
        \hline
        SHERF (Mo) & 26.95 & 44.12 & 0.9615 & 24.23 & 61.44 & 0.9450 \\
        SHERF (MV) & 26.35 & 43.68 & 0.9603 & - & - & - \\
        \hline
        GHuNeRF+ (1 obs.) & 23.89 & 44.00 & 0.9527 & 23.17 & 50.24 & 0.9480 \\
        GHuNeRF+ (2 obs.) & 23.97 & 43.72 & 0.9530 & 23.27 & 49.96 & 0.9483 \\
        GHuNeRF+ (4 obs.) & 24.00 & 43.66 & 0.9531 & 23.31 & 49.86 & 0.9485 \\
        GHuNeRF+ (8 obs.) & 24.01 & 43.64 & 0.9531 & 23.32 & 49.85 & 0.9485 \\
        \hline
        GHuNeRF (1 obs.) & 23.87 & 63.01 & 0.9474 & 23.30 & 68.84 & 0.9425 \\
        GHuNeRF (2 obs.) & 23.88 & 62.98 & 0.9474 & 23.34 & 68.76 & 0.9427 \\
        GHuNeRF (4 obs.) & 23.89 & 63.02 & 0.9474 & 23.36 & 68.76 & 0.9427 \\
        GHuNeRF (8 obs.) & 23.88 & 63.06 & 0.9474 & 23.36 & 68.75 & 0.9427 \\
        \hline
        \textbf{Ours} (1 observed) & 26.70 & 33.43 & 0.9638 & 25.08 & 42.28 & 0.9553 \\
        \textbf{Ours} (2 observed) & 27.38 & 30.20 & 0.9670 & 25.33 & 40.93 & 0.9568 \\
        \textbf{Ours} (3 observed) & 27.64 & 28.88 & 0.9683 & \textbf{25.40} & 40.53 & 0.9573 \\
        \textbf{Ours} (4 observed) & \textbf{27.66} & \textbf{28.72} & \textbf{0.9685} & \textbf{25.40} & \textbf{40.52} & \textbf{0.9574} \\
        \hline
    \end{tabular}
    \caption{Extended quantitative comparison of our method with SHERF \cite{hu_sherf_2023} and GHuNeRF \cite{li_ghunerf_2024} with various numbers of observed views on the HuMMan~\cite{cai_humman_2022} dataset. SHERF (Mo) is trained in our monocular framework, and SHERF (MV) is the official model from \cite{hu_sherf_2023} (multi-view trained). GHuNeRF+ contains the added LPIPS loss. $\textrm{LPIPS*} = \textrm{LPIPS} \times 10^3$.}
    \label{tab:quantitative-comparison-extra-humman}
\end{table*}

\begin{table*}
    \centering
    \begin{tabular}{l| c c c | c c c}
        \hline
        \multirow{2}{*}{Method} & \multicolumn{3}{c|}{\textit{Accurate body parameters}} & \multicolumn{3}{c}{\textit{Estimated body parameters}} \\
        \cline{2-7}
        & PSNR $\uparrow$ & LPIPS* $\downarrow$ & SSIM $\uparrow$ & PSNR $\uparrow$ & LPIPS* $\downarrow$ & SSIM $\uparrow$ \\
        \hline
        SHERF (Mo) & 28.49 & 48.22 & 0.9635 & 26.93 & 61.97 & 0.9536 \\
        SHERF (MV) & 27.78 & 49.52 & 0.9614 & - & - & - \\
        \hline
        GHuNeRF+ (1 obs.) & 26.59 & 53.10 & 0.9578 & 26.19 & 56.46 & 0.9547 \\
        GHuNeRF+ (2 obs.) & 26.69 & 52.92 & 0.9581 & 26.28 & 56.16 & 0.9550 \\
        GHuNeRF+ (4 obs.) & 26.70 & 52.93 & 0.9582 & 26.31 & 56.11 & 0.9552 \\
        GHuNeRF+ (8 obs.) & 26.71 & 52.94 & 0.9583 & 26.32 & 56.09 & 0.9552 \\
        \hline
        GHuNeRF (1 obs.) & 27.59 & 70.05 & 0.9562 & 27.12 & 74.74 & 0.9520 \\
        GHuNeRF (2 obs.) & 27.72 & 69.76 & 0.9566 & 27.24 & 74.59 & 0.9524 \\
        GHuNeRF (4 obs.) & 27.78 & 69.71 & 0.9568 & 27.28 & 74.54 & 0.9527 \\
        GHuNeRF (8 obs.) & 27.81 & 69.71 & 0.9569 & 27.31 & 74.58 & 0.9527 \\
        \hline
        \textbf{Ours} (1 observed) & 27.86 & 40.25 & 0.9630 & 27.00 & 47.21 & 0.9575 \\
        \textbf{Ours} (2 observed) & 28.35 & 38.03 & 0.9651 & 27.31 & 45.45 & 0.9592 \\
        \textbf{Ours} (3 observed) & 28.63 & 36.88 & 0.9663 & 27.45 & 44.79 & 0.9599 \\
        \textbf{Ours} (4 observed) & \textbf{28.65} & \textbf{36.89} & \textbf{0.9664} & \textbf{27.46} & \textbf{44.76} & \textbf{0.9601} \\
        \hline
    \end{tabular}
    \caption{Extended quantitative comparison of our method with SHERF \cite{hu_sherf_2023} and GHuNeRF \cite{li_ghunerf_2024} with various numbers of observed views on the DNA-Rendering~\cite{cheng_dna-rendering_2023} dataset. SHERF (Mo) is trained in our monocular framework, and SHERF (MV) is trained in the multi-view framework of \cite{hu_sherf_2023}. GHuNeRF+ contains the added LPIPS loss. $\textrm{LPIPS*} = \textrm{LPIPS} \times 10^3$.}
    \label{tab:quantitative-comparison-extra-dna}
\end{table*}

\section{Additional Results}
\label{sec:add-results}
Note that a fair comparison to the related GNH~\cite{masuda_generalizable_2024} is not currently possible since the code and models have not yet been made publicly available.

See \cref{tab:quantitative-comparison-extra-humman} and \cref{tab:quantitative-comparison-extra-dna} for an extended quantitative comparison to SHERF~\cite{hu_sherf_2023} (monocular -- Mo and multi-view -- MV), GHuNeRF~\cite{li_ghunerf_2024} and GHuNeRF+ on HuMMan~\cite{cai_humman_2022} and DNA-Rendering~\cite{cheng_dna-rendering_2023}. SHERF (MV) is trained in the original framework of \cite{hu_sherf_2023}, \ie the observed view is in the same pose as the target view but captured from a different viewpoint. Note that SHERF (MV) is still conditioned on a single observed view. For `SHERF (MV)` on the HuMMan dataset we use the official models of \cite{hu_sherf_2023}, while for the DNA-Rendering dataset we retrain it using the multi-view training framework.

\begin{figure*}
    \centering
    \includegraphics[width=\textwidth,height=0.97\textheight,keepaspectratio]{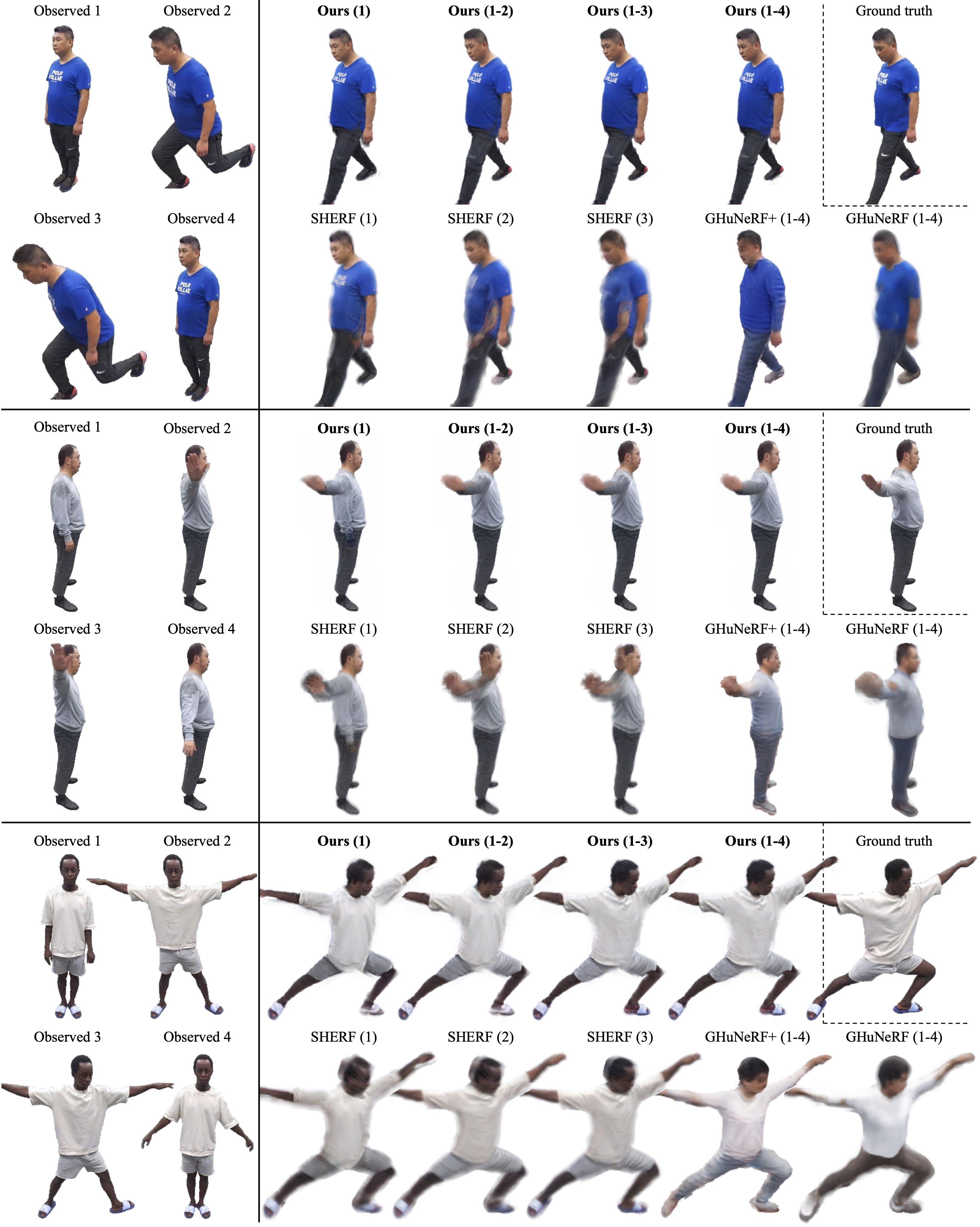}
    \caption{Extended qualitative comparison between our method, SHERF (Mo), and GHuNeRF on the HuMMan dataset. Numbers in parentheses indicate the range of observed views supplied to the respective models. Best viewed in color and zoomed in for details.}
    \label{fig:add-quali-humman}
\end{figure*}

\begin{figure*}
    \centering
    \includegraphics[width=\textwidth,height=0.97\textheight,keepaspectratio]{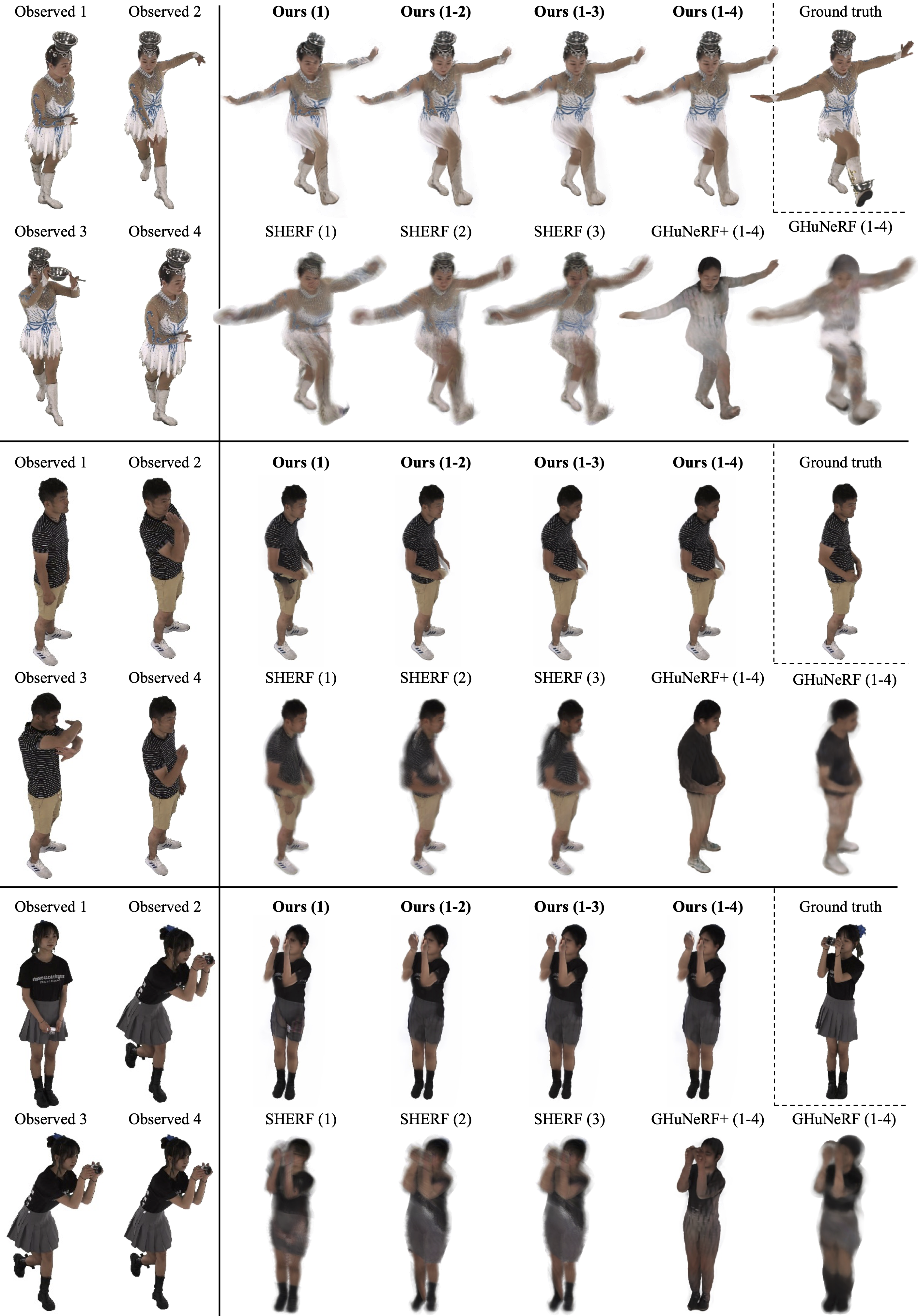}
    \caption{Extended qualitative comparison between our method, SHERF (Mo), and GHuNeRF on the DNA-Rendering dataset. Numbers in parentheses indicate the range of observed views supplied to the respective models. Best viewed in color and zoomed in for details.}
    \label{fig:add-quali-dna}
\end{figure*}

\cref{fig:add-quali-humman} and \cref{fig:add-quali-dna} show an extended qualitative comparison between our method with $T \in \{ 1, 2, 3, 4 \}$ observed views, SHERF~\cite{hu_sherf_2023} (Mo) and GHuNeRF~\cite{li_ghunerf_2024} on the HuMMan~\cite{cai_humman_2022} and DNA-Rendering~\cite{cheng_dna-rendering_2023} datasets, respectively.
As discussed in \cref{sec:quali-results}, SHERF frequently struggles to match the observed view to the underlying geometry, which results in incorrect renders in novel poses with `phantom` limbs (typically arms) imprinted on the torso (see the top 2 subjects in \cref{fig:add-quali-humman} and top two subjects in \cref{fig:add-quali-dna}). In most cases, this problem is observed regardless of which view SHERF observes -- as long as the arms of the subject overlap with their body in the observed view, they are usually imprinted somewhere on the torso.
While our method sometimes displays a similar pattern when it observes a single view, it matches the geometry correctly and resolves this issue when provided with 2 (or more) observations. To achieve that, it has to combine information from available observations while resolving occlusions and/or making use of the prior (\eg smoothness), as information from any of the observations alone is not enough to eliminate the artifacts (which is demonstrated by SHERF results). 

\subsection{Extended Results with Estimated Body Shape and Pose Parameters}
\label{sec:add-results-estim}
\begin{figure*}
    \centering
    \includegraphics[width=\textwidth,height=0.96\textheight,keepaspectratio]{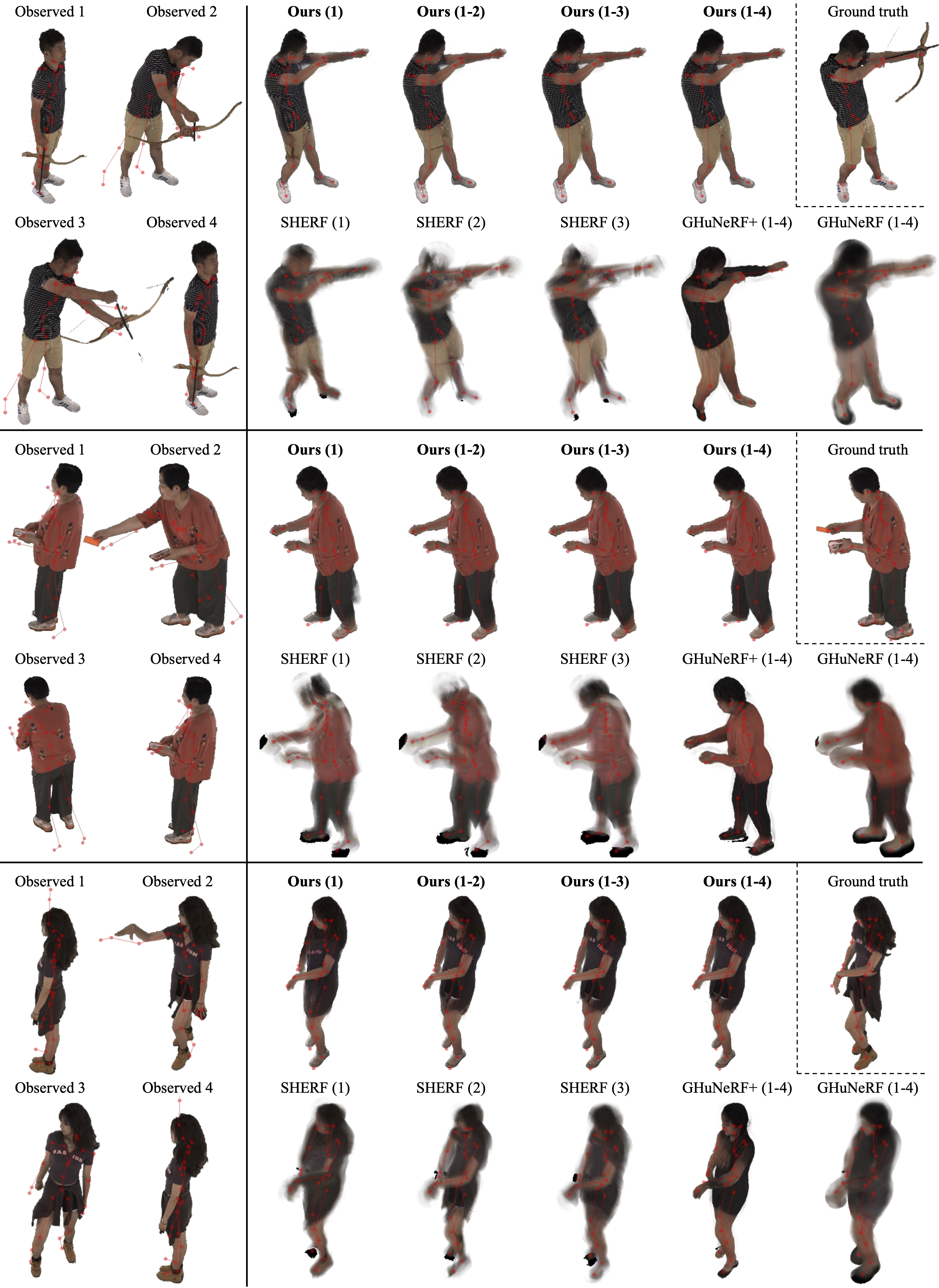}
    \caption{Extended qualitative comparison between our method, SHERF (Mo), and GHuNeRF on the DNA-Rendering dataset when using estimated body shape and pose parameters. Numbers in parentheses indicate the range of observed views supplied to the respective models. Best viewed in color and zoomed in for details.}
    \label{fig:add-quali-dna-estim}
\end{figure*}

\cref{fig:add-quali-humman-estim} and \cref{fig:add-quali-dna-estim} show an extended qualitative comparison of our method with $T \in \{ 1, 2, 3, 4 \}$ observed views to SHERF (Mo) and GHuNeRF, on the HuMMan and DNA-Rendering datasets (respectively) when using estimated body shape and pose parameters.
The renders produced by our method are significantly sharper compared to SHERF and, in contrast to the baselines, our method correctly replicates most of the details found in the observed views. Moreover, our method generates fewer artifacts compared to SHERF when filling in missing information using prior (see \eg the legs and shoes of all subjects in \cref{fig:add-quali-dna-estim}).

\subsection{Video Qualitative Results}
\begin{figure*}
    \centering
    \includegraphics[width=\textwidth,height=0.96\textheight,keepaspectratio]{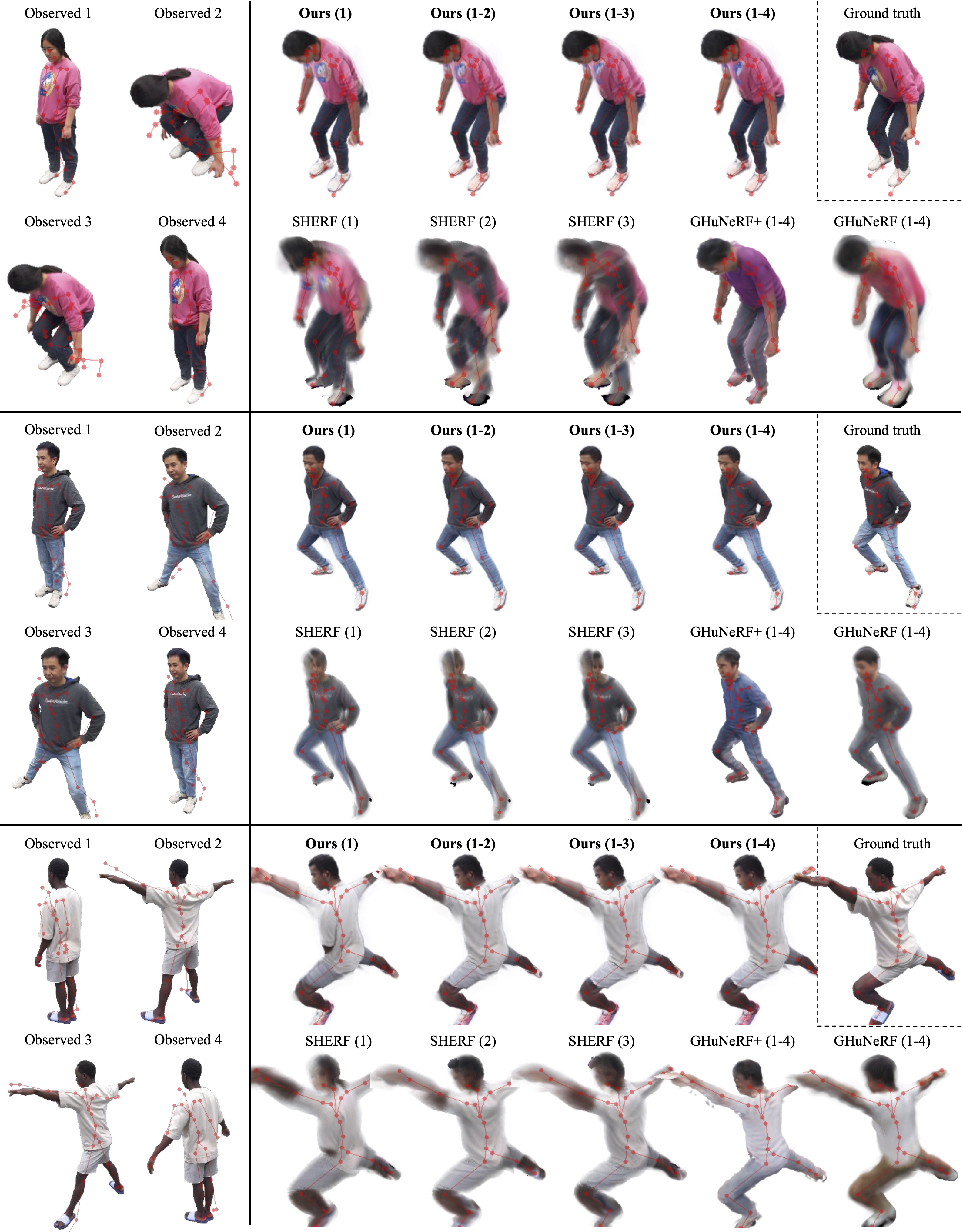}
    \caption{Extended qualitative comparison between our method, SHERF (Mo), and GHuNeRF on the HuMMan dataset when using estimated body shape and pose parameters. Numbers in parentheses indicate the range of observed views supplied to the respective models. Best viewed in color and zoomed in for details.}
    \label{fig:add-quali-humman-estim}
\end{figure*}

Please see the project page for video versions of \cref{fig:add-quali-humman}, \cref{fig:add-quali-dna}, \cref{fig:add-quali-humman-estim} and \cref{fig:add-quali-dna-estim}.

\section{Broader impact}
We acknowledge that our method could potentially have a negative societal impact if misused to create fake images or videos of real people. Any public deployments of this technology should be done with great care to ensure that ethical guidelines are met and with safeguards in place. We will release our code publicly to aid with countermeasure analysis.

